\documentclass[submission,copyright,creativecommons]{eptcs}

\makeatletter
\let\O@argtabularcr\@argtabularcr
\def\O@xtabularcr{\@ifnextchar[\O@argtabularcr{\ifnum 0=`{\fi}\cr}}
\let\O@tabacol\@tabacol
\let\O@tabclassiv\@tabclassiv
\let\O@tabclassz\@tabclassz
\let\O@tabarray\@tabarray
\def\author@tabular{\authorsize\def\@halignto{}\@authortable}
\let\endauthor@tabular=\endtabular
\def\author@tabcrone{{\ifnum0=`}\fi\O@xtabularcr\affilsize\itshape
 \let\\=\author@tabcrtwo\ignorespaces}
\def\author@tabcrtwo{{\ifnum0=`}\fi\O@xtabularcr[-3\p@]\affilsize\itshape
 \let\\=\author@tabcrtwo\ignorespaces}
\def\@authortable{\leavevmode \hbox \bgroup $\let\@acol\O@tabacol
 \let\@classz\O@tabclassz \let\@classiv\O@tabclassiv
 \let\\=\author@tabcrone \ignorespaces \O@tabarray}
\makeatother

\providecommand{\event}{ICLP 2020} 

\usepackage[table,xcdraw]{xcolor}
\usepackage[fleqn]{amsmath}
\usepackage{algorithm}
\usepackage{algpseudocode}
\usepackage[inline]{enumitem}
\usepackage{graphicx}
\usepackage{tikz-cd}
\usepackage{multirow} 
\usepackage{cprotect}
\usepackage{comment}
\usepackage{caption}
\usepackage{diagbox}
\usepackage{listings}
\usepackage{wrapfig}
\usepackage{fixltx2e}
\usepackage[subtle]{savetrees}
\usepackage[compact]{titlesec}
\usepackage[font=small,skip=0pt]{caption}
\definecolor{darkbrown}{rgb}{0.4, 0.26, 0.13}
\definecolor{burgundy}{rgb}{0.5, 0.0, 0.13}

\titlespacing{\section}{2.75pt}{2.75pt}{2.75pt}
\titlespacing{\subsection}{2pt}{2pt}{2pt}

\title{SQuARE: Semantics-based Question Answering and Reasoning Engine\thanks{Work partially supported by NSF awards IIS 1718945, IIS 1910131, IIP 1916206, and a DARPA award.}}
\author {Kinjal Basu, Sarat Chandra Varanasi, Farhad Shakerin and Gopal Gupta\\
         The University of Texas at Dallas, Richardson, USA\\
         \email{\scriptsize \{Kinjal.Basu, Sarat-Chandra.Varanasi, Farhad.Shakerin, Gopal.Gupta\}@utdallas.edu}}
         
\def\titlerunning{Semantics-based Question Answering and Reasoning Engine}
\def\authorrunning{Basu et al.}
\begin{document}
\maketitle

\begin{abstract}
Understanding the meaning of a text is a fundamental challenge of natural language understanding (NLU) and from its early days, it has received significant attention through question answering (QA) tasks. We introduce a general semantics-based framework for natural language QA and also describe the SQuARE system, an application of this framework. The framework is based on the denotational semantics approach widely used in programming language research. In our framework, valuation function maps syntax tree of the text to its commonsense meaning represented using basic knowledge primitives (the semantic algebra) coded using answer set programming (ASP). We illustrate an application of this framework by using VerbNet primitives as our semantic algebra and a novel algorithm based on partial tree matching  that generates an answer set program that represents the knowledge in the text.  A question posed against that text is converted into an ASP query using the same framework and executed using the s(CASP) goal-directed ASP system. Our approach is based purely on (commonsense) reasoning. SQuARE achieves 100\% accuracy on all the five datasets of bAbI QA tasks that we have tested. The significance of our work is that, unlike other machine learning based approaches, ours is based on ``understanding'' the text and does not require any training. SQuARE can also generate an explanation for an answer while maintaining high accuracy. 
\end{abstract}


\section{Introduction}


Semantic representation of natural language is a long term goal of natural language understanding (NLU) research. To understand the meaning of a natural language sentence, humans first process the syntactic structure of the sentence and then infer the meaning next. Also, humans use commonsense knowledge to understand the often complex and ambiguous meaning of natural language sentences. 
Humans interpret a passage as a sequence of sentences and will normally process the events in the story in the same order as the sentences. Once humans understand the meaning of a passage, they can answer questions posed, along with an explanation for the answer. We believe that an automated question answering (QA) system should work in a similar way. 
Thus, an ideal automated QA framework should map each sentence to the knowledge it represents, then augment it with commonsense knowledge related to the concepts involved (e.g., if the passage involves a lion, the system should know that a lion is a living being, an animal, carnivore, etc.), then use the combined knowledge to automatically answer questions. Questions should be first automatically converted into a query that is executed against this knowledge. We develop such a framework in this paper that relies on denotational semantics to specify the mapping from syntax to knowledge (meaning). This denotational, automated QA framework has been employed to realize the SQuARE engine described in this paper.

Researchers have created several QA datasets to test a QA system, such as SQuAD-1.0 \cite{squad1}, bAbI \cite{babi}, MCTest \cite{mctest}, SQuAD-2.0 \cite{squad2}, WikiQA \cite{wikiqa}, etc. However, most of the state-of-the-art QA systems have been developed using machine learning (ML) techniques. The common drawback among all of these ML-based systems is that they answer questions by exploiting the correlation among words and without properly understanding the content. These systems are also not explainable, i.e., cannot generate a justification of their conclusions. They may perform well for a particular dataset or a domain, but will fail on other tasks. Humans can easily generalize from one type of question to another, even if they have not encountered it before. This leads to another problem for all these machine learning based systems: they fail to answer questions for which they have not seen training data. Most of such ML-based QA systems work as information retrieval agents that can answer questions against a textual passage with either shallow (or, no) reasoning capabilities. In this paper, we present a general semantics-based  framework for automated QA that is a step towards mimicking the human question answering mechanism. 

We also introduce our novel algorithm for automatically generating the knowledge (semantics) corresponding to each English sentence. The algorithm leverages the comprehensive verb-lexicon for English verbs---VerbNet \cite{vn}. For each English verb, VerbNet gives the syntactic and semantic patterns. 
The algorithm
employs partial syntactic matching between parse-tree of a sentence and a verb's \textit{frame syntax} from VerbNet to obtain the meaning of the sentence in terms of VerbNet's primitive predicates. This matching is motivated by denotational semantics and can be thought of as mapping parse-trees of sentences to knowledge that is constructed out of semantics provided by VerbNet. The VerbNet semantics is expressed using a set of primitive predicates that can be thought of as the \textit{semantic algebra} of the denotational semantics \cite{d_semantics}.

SQuARE has been tested on five datasets (\textit{single supporting fact}, \textit{two supporting facts}, \textit{three supporting facts}, \textit{two argument relations} and \textit{positional reasoning}) of the bAbI QA tasks, that is created by Facebook-Research for testing text understanding and reasoning of a QA system. The bAbI QA datasets consist of stories. Each story is a collection of simple sentences and the questions are asked in between a set of sentences interactively or at the end of the story. Answer for each question can be computed by performing commonsense reasoning over the information captured from the earlier sentences. We choose bAbI QA dataset to test SQuARE, as the sentences involved are simpler and require less amount of pre-processing work so we can concentrate on the semantic representation and reasoning tasks. The SQuARE system achieves 100\% accuracy on all five tasks tried, outperforming state-of-the-art systems.  It also surpasses other systems wrt explainability as it can provide proper justification for each answer.

This paper makes the following novel contributions: (i) furnishes a fully explainable semantic-based general framework for textual question answering, (ii) demonstrates that a QA application---SQuARE---developed using the framework can  reach 100\% accuracy without training; it also outperforms the machine learning based models in terms of explainability, (iii) showcases the novel partial tree matching algorithm using VerbNet primitives as semantic algebra for the generation of a sentence semantics, (iv) shows how commonsense reasoning can be included for QA in dynamic domains, and (v) provides a method that relies on goal-directed ASP that guarantees a correct answer as long as the semantic representation and the query generation are correct. Our work is purely based on reasoning and does not require any manual intervention other than providing reusable commonsense knowledge coded in ASP. It paves the way for developing advanced NLU systems based on ``truly understanding'' the text. 

\section{Background}\label{background}

We next describe some of the key technologies we employ.
Our effort is based on answer set programming (ASP) technology \cite{stablemodel,asp}, specifically, its goal-directed implementation in the s(ASP) and s(CASP) systems \cite{sasp,scasp}. The semantics of a sentence and \textit{frame semantics} of VerbNet that we use are represented as answer set programs. The use of a goal-directed implementation of ASP is critical to our project. We assume that the reader is familiar with ASP. Expositions of ASP can be found elsewhere \cite{asp}. 

\smallskip
\noindent\textbf{The s(ASP) System:}
    s(ASP) \cite{sasp}  is a query-driven, goal-directed implementation of ASP. The s(CASP) system \cite{scasp} is an extension of s(ASP) that adds constraint solving over reals. Goal-directed execution of s(ASP)/s(CASP) is indispensable for automating commonsense reasoning, as traditional grounding and SAT-solver based implementations of ASP may not be scalable. There are three major advantages of using s(ASP) (and s(CASP)) system: 
    \begin{enumerate*}
        \item[(i)] s(ASP) does not ground the program, which makes our framework scalable,
        \item[(ii)] it only explores the parts of the knowledge base that are needed to answer a query, and
        \item[(iii)] it provides justification  (proof tree) for an answer.
    \end{enumerate*}

\smallskip

\noindent\textbf{Stanford CoreNLP Tool Set:}
Stanford CoreNLP \cite{corenlp}  is a set of tools for natural language processing. SQuARE only uses \textit{parts-of-speech (POS) tagger}, \textit{named-entity recognizer (NER) tagger}, \textit{constituency parser} and \textit{dependency parser} from this set. Based on the context, POS tagger generates the necessary parts
\begin{wrapfigure}{r}{0.48\linewidth}
\includegraphics[scale=0.6]{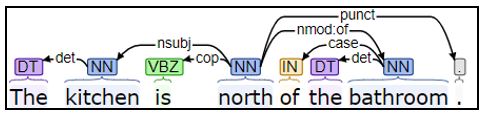}
\caption{{Example of POS tagging and dependency graph}}
\label{fig:dependency_graph}
\end{wrapfigure}
   of speech such as noun, verb, adjective, etc., for each statement and also identifies the question type (e.g., what, where, who). The NER tagger identifies named-entities, such as a person's name. Constituency parser produces the syntactic parse tree (shown in figure \ref{fig:parse-tree}) of a sentence where the node's value follows Penn-Treebank's tags \cite{penntree}. The dependency graph provides the grammatical relations between words in a sentence. Dependency relations follow enhanced universal dependency annotation \cite{uni_dependency}. Figure \ref{fig:dependency_graph} shows an example of the POS tagging and the dependency graph of a statement from the  \textit{two argument relations} task from the bAbI dataset \cite{babi}.

\noindent

\begin{figure}[b]
\scriptsize
\centering
\begin{tabular}{|p{1.5cm}|p{11.5cm}|}
    \hline
     \multicolumn{2}{|l|}{\textbf{NP V NP PP.Destinations}}   \\
    \hline
    \textbf{Example}  & \textbf{\textit{`` Steve tossed the ball to the garden.''}} \\
    \hline 
    \textbf{Syntax}  & \textbf{Agent} \textbf{V} \textbf{Theme} \textcolor{blue}{\{\{+}\textcolor{red}{Dest}| \textcolor{blue}{+}\textcolor{red}{Loc}\textcolor{blue}{\}\}} \textbf{Destination} \\
    \hline  
    \textbf{Semantics}  & \textcolor{burgundy}{Exert\_Force(}\textbf{During(E\textsubscript{0}),Agent,Theme}\textcolor{burgundy}{)},  \textcolor{burgundy}{Contact(}\textbf{End(E\textsubscript{0}),Agent,Theme}\textcolor{burgundy}{)}, \\
    & \textcolor{burgundy}{Motion(}\textbf{During(E\textsubscript{1}),Theme}\textcolor{burgundy}{)}, \textcolor{red}{$\neg$} \textcolor{burgundy}{Contact(}\textbf{During(E\textsubscript{1}), Agent, Theme}\textcolor{burgundy}{)},  \\
    &  \textcolor{burgundy}{Path\_Rel(}\textbf{Start(E\textsubscript{0}), Theme, Initial\_Location, Ch\_of\_Loc, Prep}\textcolor{burgundy}{)}, \\
    & \textcolor{burgundy}{Path\_Rel(}\textbf{End(E\textsubscript{1}), Theme, Destination, Ch\_of\_Loc, Prep}\textcolor{burgundy}{)}, \\
    & \textcolor{burgundy}{Cause(}\textbf{Agent, E\textsubscript{1}}\textcolor{burgundy}{)}, \textcolor{burgundy}{Meets(}\textbf{E\textsubscript{0}, E\textsubscript{1}}\textcolor{burgundy}{)}, \textcolor{burgundy}{Equals(}\textbf{Agent, Initial\_Location}\textcolor{burgundy}{)} \\ 
    \hline
\end{tabular}
\vspace{0.02in}
\caption{VerbNet frame instance for the verb class \textit{discard}} 
    \label{fig:verbnet-example}
\end{figure}

\smallskip
\noindent\textbf{VerbNet:} 
VerbNet \cite{vn} is the largest online network of English verbs. It is inspired by Beth Levin’s classification of verb classes and their syntactic alternations \cite{verbclasses}. A verb class in VerbNet is mainly expressed by \textit{syntactic frames}, \textit{thematic roles},  and  \textit{semantic representation}. The VerbNet
lexicon identifies  thematic roles and  syntactic patterns of each verb class and infers the common syntactic structure and semantic relations for all the member verbs. Figure \ref{fig:verbnet-example}  shows an example of a VerbNet frame of the verb class \textit{discard}. Note that there can be multiple frames for a given verb class, one for each type of usage  of the verb.
The first line in the figure (\textit{NP V NP PP.Destinations}) shows the syntax tree pattern for this particular usage. The second line shows an example of English sentence that matches the usage. The third line  (we call it \textit{frame syntax}) shows the constraints on the syntactic categories NP, PP, etc. (e.g., the first noun phrase should be of type AGENT). The final line (we call it \textit{frame semantics}) shows the meaning expressed using basic primitive predicates of VerbNet (\textit{exert\_force, contact, motion}, etc.). A verb in an English sentence generally corresponds to an event. The meaning expressed using primitive predicates represents the semantic relations between event participants across the course of the event.

\section{General Architecture for Semantics-based NLP QA}
Denotational semantics has been extensively used in the study of programming languages. Broadly, denotational semantics is concerned with defining the meaning of a program in terms of mathematical objects called domains, such as integers, truth values, tuples of values, and mathematical functions \cite{d_semantics}. 
Typically, there are three components of denotational semantics of a programming language: 
\begin{itemize}
    \item \textit{Syntax}: specified as abstract syntax trees.
    \item \textit{Semantic Algebra}: these are the basic domains along with the associated operations; meaning of a program is expressed in terms of these basic domains.
    \item \textit{Valuation Function}: these are mappings from abstract syntax trees and semantic algebra to values in the  domains in the semantic algebra.
\end{itemize}

\begin{figure*}[h]
    \centering
    \includegraphics[width = 13cm]{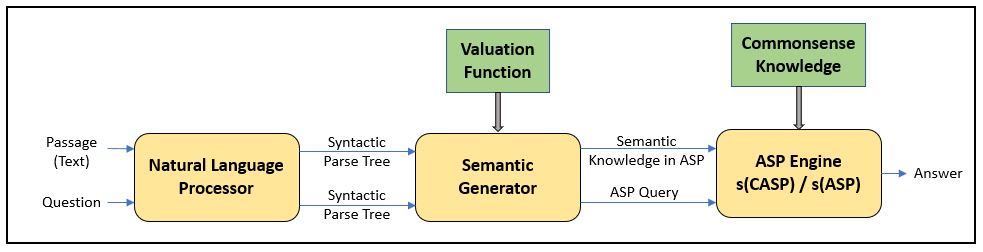}
    \caption{ General Framework for Semantic-based QA System }
    \label{fig:genral_framework}
    \end{figure*}

To ``truly understand'' a text, an ideal NLU system should transform the natural language text into knowledge expressed using the semantic algebra of well-understood concepts. Then, this semantics can be used for different NLU tasks like question answering, summarization, information extraction, etc. With the formalized textual information, the semantics, and commonsense knowledge, an NLU system can easily perform basic reasoning such as counting, comparing, inference, spatial reasoning. Figure \ref{fig:genral_framework} illustrates a general framework for textual question answering and is explained next.

\noindent

\noindent \textbf{Natural Language Processor}: performs the grammatical (syntax) analysis (e.g., lemmatization, parts-of-speech tagging, parsing, etc.) of a natural language sentence and generates the syntactic parse tree via constituency parsing. Many PCFG (probabilistic context free grammar) or deep learning based state-of-the-art parsers are available that find the most probable parse tree from all the trees generated due to the ambiguous nature of the English language.

\noindent \textbf{Valuation Function}: takes the parse tree as an input and maps it to the knowledge entailed in the sentence. This knowledge is represented in terms of knowledge primitives (semantic algebra) whose meaning is well-understood.

\noindent \textbf{Semantics Generator}:
The semantics generator applies the valuation function to parse trees using the partial tree matching algorithm to generate the knowledge encapsulated in a sentence. This knowledge is represented as an answer set program. \textit{Partial} tree matching is needed due to the complexity and ambiguity of natural languages.

\noindent \textbf{Commonsense Knowledge}: This module represents the background commonsense knowledge needed for reasoning. When some information is not explicitly mentioned in a passage and the system needs it to make inferences, this commonsense knowledge becomes crucial. 
The commonsense knowledge is also coded in ASP for consistency of notation. Also, our inference algorithms require that all knowledge be coded in ASP.
This handcrafted commonsense knowledge has been represented in a generic manner and can be reused in future. This method is very similar to a human knowledge gathering process. We acquire/refine our commonsense knowledge incrementally from information we encounter in our everyday life and we use it for reasoning later when needed.
Commonsense knowledge can be thought of as augmenting the semantic algebra of knowledge primitives. It is also employed to augment knowledge output by valuation functions. Note that all commonsense knowledge is generally represented as rules that define defaults,  exceptions and preferences coded in ASP. Our goal is to keep the inference algorithm---default reasoning with exceptions and preferences---as simple as possible. This simplicity is important as otherwise it is possible that our encoding contains the knowledge needed to answer a question but we are not able to answer it because of the complexity of our decision procedures and not knowing which decision procedure to apply and when.

\noindent \textbf{s(CASP) ASP Engine}: The s(CASP) goal-directed ASP engine executes the query generated from the question against the knowledge generated from the text. The s(CASP) system performs a goal-directed search of the knowledge base while applying defaults, resolving exceptions, and retracting defeasible worlds. The (partial) stable models produced in response to the query represents the answer to the question. Moreover, s(CASP) performs all these tasks without grounding the program, avoiding exponential blowup in the grounded program size common in SAT-based systems.  


\section{Semantics-driven QA: the SQuARE System} \label{section:semantic}

Unlike programming languages, the denotation of a natural language can be quite ambiguous. English is no exception and the meaning of a word or sentence may depend on the context. The generation of correct knowledge from a sentence, hence, is quite hard. We next present a novel approach to automatically mapping parse trees of simple English sentences to their denotations, i.e., knowledge they represent.

\noindent
\begin{figure*}[t]
    \centering
    \includegraphics[width = 13cm]{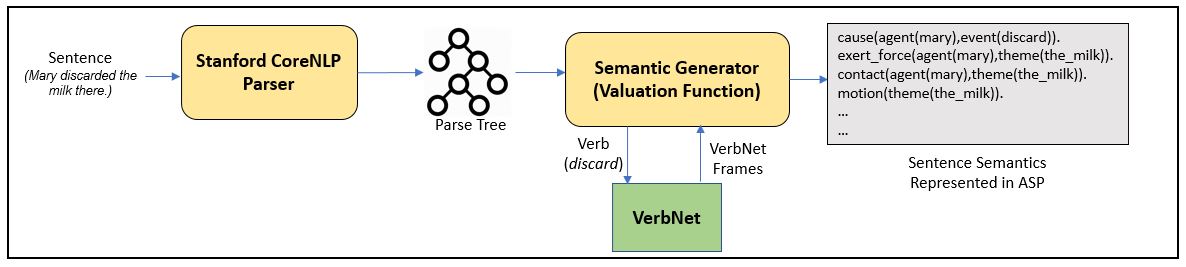}
    \caption{ Sentence to semantic generation process in SQuARE }
    \label{fig:semantic_algebra}
    \end{figure*}

\subsection{Semantics-driven translation of English sentences}\label{section:semantic_algebra}

\setlength{\textfloatsep}{6pt}
\begin{algorithm}[b]
    \small
    \caption{Semantic Knowledge Generation}\label{algorithm1}
    \hspace*{\algorithmicindent} \textbf{Input:} \textit{$p_t$}: \text{constituency parse tree of a sentence}\\ 
    \hspace*{\algorithmicindent} \textbf{Output:} \textit{semantics}: \text{sentence semantics} 
    \begin{algorithmic}[1]
        \Procedure{GetSentenceSemantics}{$p_t$}
        \State $ \textit{verbs} ~ \gets  ~ \textit{getVerbs($p_t$)}$ \Comment{returns list of verbs present in the sentence}
        \State $ \textit{semantics} ~  \gets ~ \textit{\{\}}$ \Comment{initialization}
        \State \textbf{for each} {$v ~ \in ~ \textit{verbs}$} \textbf{do}
            \State \hspace{4mm} $ \textit{classes} ~ \gets ~ \textit{getVNClasses(v)}$ \Comment{get the VerbNet classes of the verb}
            \State \hspace{4mm} \textbf{for each} {$c ~ \in ~ \textit{classes}$} \textbf{do}
                \State \hspace{10mm}$ \textit{frames} ~ \gets ~ \textit{getVNFrames(c)}$ \Comment{get the VerbNet frames of the class}
                \State \hspace{10mm} \textbf{for each} {$f ~ \in ~ \textit{frames}$} \textbf{do}
                    \State \hspace{15mm} $ \textit{thematicRoles} ~ \gets ~ \textit{getThematicRoles($p_t$, f.syntax, v)}$  \Comment{see Algorithm \ref{algorithm2}}
                    \State \hspace{15mm} $ \textit{semantics} ~ \gets ~ \textit{semantics} \; \; \;  \cup \; \; \;  \textit{getSemantics(thematicRoles, f.semantics)}$ \\ \Comment{map the thematic roles into the frame semantics}
                \State \hspace{10mm} \textbf{end for}
            \State \hspace{4mm} \textbf{end for}
        \State \textbf{end for}
        \State \Return{semantics}
        \EndProcedure
    \end{algorithmic}
\end{algorithm}
The verb plays a significant role in an English sentence as it represents the event encapsulated in the sentence. The verb also constrains the relation among event participants. As described earlier (Section \ref{background}), the VerbNet system captures all of this information through
 verb classes that represent a group of verbs that share common meaning (e.g., toss and discard). As discussed earlier, VerbNet also provides a skeletal parse tree (frame syntax) for different usages of a verb and the corresponding semantics (frame semantics) that represent the knowledge. Thus, VerbNet can be thought of as a very large valuation function that maps syntax tree patterns to their respective meanings: essentially, a giant case statement represented through the large collection of VerbNet frames, where each frame represents the mapping of one specific type of usage of a verb class to its meaning. 
 Thus, in Fig. \ref{fig:verbnet-example},  knowledge contained in the sentence ``Steve tossed the ball to the garden.'' is represented by the conjunction of predicates in the frame semantics (line 4), where predicate argument AGENT is instantiated to `Steve', THEME to `the ball', DESTINATION to 'the garden', etc.  Fig. \ref{fig:semantic_algebra} shows another example.

Thus, given the parse tree of a sentence, $p_t$, the SQuARE system matches VerbNet's syntax tree skeletons of the verbs occurring in the sentence with sub-trees of $p_t$. The predicates corresponding to the matching skeletons are assembled together to produce the knowledge represented in the sentence, as discussed earlier and
illustrated in figure \ref{fig:semantic_algebra}. The knowledge is represented in terms of VerbNet primitive predicates (semantic algebra).
For example, one relation in the semantic algebra of the verb \textit{discard} is \textit{CAUSE}. The relation \textit{CAUSE} is a binary relation qualifying the actor of \textit{discard} with the verb \textit{discard} itself. That is, \textit{CAUSE} is represented as\textit{ CAUSE(Agent, event(discard))} where \textit{Agent} is a thematic role from VerbNet.

\begin{wrapfigure}{r}{0.33\linewidth}
\includegraphics[scale=0.7]{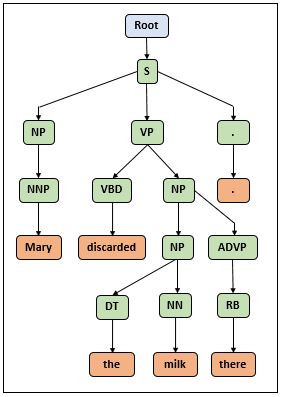}
\caption{{Parse-Tree}}
\label{fig:parse-tree}
\end{wrapfigure}

In the SQuARE system, first, the English sentence is parsed using Stanford's CoreNLP parser to obtain the syntactic parse tree of the sentence. In the parse tree (illustrated in Figure \ref{fig:parse-tree}), there will always be a root node (blue colored), named \textit{Root}, from where the tree expands. All the internal nodes (green colored) possess the \textit{word, phrase} or \textit{clause} tags from Penn-Treebank \cite{penntree}. The word-level tags are the parts-of-speech tag of each word, such as NN (nominal noun), VBD (verb: past tense), etc. The example of phrase-level tags are NP \textit{(noun phrase)}, VP \textit{(verb phrase)}, PP \textit{(preposition phrase)}, etc., and clause-level tags are S \textit{(simple declarative clause)}, SBARQ \textit{(Direct question introduced by a wh-word or a wh-phrase)}, etc. The leaf nodes (orange colored) of the parse tree are the words from the sentence with the parts-of-speech node as their parent. Figure \ref{fig:parse-tree} shows the syntactic parse tree of the sentence -\textit{``Mary discarded the milk there.''}. 

Next, the generated syntactic parse tree of a sentence is passed as an input to our \textit{Semantic Knowledge Generation} algorithm \ref{algorithm1} that maps the sentence to its meaning. To accomplish this, the list of verbs mentioned in the sentence are collected and their syntactic and semantic information is determined. As discussed earlier, each verb is a part of one or more VerbNet classes and each class has multiple frames. A frame provides information about the syntactic structure of the verb along with thematic roles. The semantic definition of each frame uses  pre-defined predicates of VerbNet that have
thematic-roles (AGENT, THEME, etc.) as arguments. These predicates are represented in ASP. \textit{Our goal is to find the partial matching between the parse tree and the frame's syntactic structure and ground the thematic-role variables so that we can get the semantics of the sentence from the frame semantics and represent it in ASP.}

\setlength{\textfloatsep}{6pt}
\begin{algorithm}[b]
    \small
    \caption{Partial Tree Matching}\label{algorithm2}
    \hspace*{\algorithmicindent} \textbf{Input:} \textit{$p_t$}: \text{constituency parse tree of a sentence}; \textit{s}: \text{frame syntax}; \textit{v}: \text{verb}\\
    \hspace*{\algorithmicindent} \textbf{Output:} \textit{tr}: \text{thematic role set} or \textit{empty-set}: \text{\{\}} 
    \begin{algorithmic}[1]
        \Procedure{GetThematicRoles}{$p_t$, s, v}
        \State $ \textit{root} ~ \gets ~ \textit{getSubTree(node(v), $p_t$)}$  \Comment{returns the sub-tree from the parent of the verb node}
        \While{root}
           \State $tr ~ \gets ~ getMatching(root, s)$  \Comment{if s matches the tree return thematic-roles, else \{\}} 
           \If {$tr ~\neq ~\textit{\{\}}$} \Return tr  
            \EndIf
            \State $ \textit{root} ~ \gets ~  \textit{getSubTree(root, $p_t$)}$  \Comment{returns \textit{false} if root equals $p_t$}

        \EndWhile
        
        \State \Return{\{\}}
        \EndProcedure
    \end{algorithmic}
\end{algorithm}

The syntax of a frame consists of word- and phrase-level tags from  the Penn Treebank, described earlier. For each frame, VerbNet also provides the mapping between the thematic roles and the tags (Figure \ref{fig:verbnet-example} illustrates an example). The algorithm starts from the verb node, performs a bottom-up search and finds the matching frame at each level through depth-first traversal. The matching mechanism can also skip words at a level if it does not fit the VerbNet frame. For instance, Algorithm 2 ignores prepositions if it is not used in the frame syntax. If an exact or partial match is found, the algorithm returns the thematic roles to the parent Algorithm \ref{algorithm1}. If a match is not found then the algorithm looks for a match at the parent level. The algorithm terminates when we reach the root without a match. In such a case, we return an empty set. This partial tree matching is summarized in Algorithm \ref{algorithm2}.

\subsection{Passage translation}
A passage or a story is a collection of sentences. Humans formulate the meaning of a passage by combining the meaning of each individual sentence. In our mind, we can quickly order the events that occur in the passage and with our commonsense knowledge, we can relate the participants involved (e.g., actor, recipient, or object) in the state before, after, and during each event. When reading a passage, humans construct the knowledge (semantics) of each sentence. The knowledge constructed is non-monotonic, for example, containing defaults and exceptions. The knowledge accumulated as we read the sentences in the passage captures the information of the events in the passage until the last sentence  read. For example, when processing the sentence - \textit{``John moved to the bathroom''}, a human would assume that \textit{John} would be at \textit{the bathroom} at the next instant, unless we encounter an exception in the next sentence, such as \textit{``But the door was locked.''}.
In other words, we arrange events in a passage in the temporal order corresponding to the order of the sentences containing these events. Predicates associated with events thus will have to have a timestamp argument to capture this ordering. We assume the events in the story have occurred sequentially unless we find exceptions. 
The meaning of a passage is composed of the meaning of each of its individual sentences. Each sentence is assigned a unique identifier. The meaning of each sentence is represented as an answer set program. The meaning of a passage is the union of the answer set programs of its sentences. 

\subsection{Commonsense Concepts}

Similar to humans, SQuARE needs to use commonsense knowledge to understand the environments and the relationships between objects and persons in the story. Each verb will have associated commonsense knowledge. As an example, consider the verb \textit{possess} arising in the bAbI dataset. SQuARE will have to compute, for example, whether an object is possessed by someone at some point in time and also compute the location of the objects from the knowledge of events in the passage. The property \textit{possession} captures the ownership of an object by a person of the story. Similarly, the property \textit{location} captures the position or place of a person in the story. 
The commonsense meaning of \textit{possess} is described below as a default rule in ASP.

\smallskip

\noindent
        \cprotect \fbox{
        \centering
        \begin{minipage}{0.96\linewidth}
            {\small
            \begin{verbatim}
property(possession, T, Per, Obj) :- path_rel(T, start(X), source(_), theme(Obj)),  
  cause(T, agent(Per), event(X)), not ab(possess_verb(X)), not unknown_value(Obj).                 \end{verbatim}
            
            }
        \end{minipage}
        }
        
\smallskip
\noindent
The \textit{path\_rel} or \textit{path-relation} predicate is part of the VerbNet semantic algebra that depicts the relationship between the start of an event/action, the \textit{source}, and the \textit{theme}. From the definition of the \textit{path\_rel} predicate, the relationship tells that the starting of event \textit{X} moves the \textit{theme} from the \textit{source} (which we do not care about now as it is unknown). \textit{Cause} predicate illustrates that in the same timestamp of the \textit{path\_rel}, the \textit{agent} causes the event. The rule has also a weak exception \cite{asp} that falsifies \textit{possession} if there is any abnormality present about the possessive verb \textit{X}. Also, the rule confirms the variable \textit{object} is not \textit{unknown} through the negated call \texttt{not unknown\_value(Obj)}. 

Commonsense law of inertia is also needed for the verb \textit{possess}. 
%
Thus, the system assumes that the object/person stays in the same state where it was in the previous time-stamp if there is no evidence about the state change of that object/person in the current time-stamp. The reverse is also true, that is, if an object is not possessed in an earlier timestamp, it is not possessed by any person in the current time-stamp, unless there is evidence to the contrary. These default rules are given below (their meaning is fairly obvious and is omitted due to lack of space; note that after(T2,T1) stands for time T2 $>$ time T1):

\smallskip
\noindent
       \cprotect \fbox{
        \centering
        \begin{minipage}{0.97\linewidth}
            {
            \small
            \begin{verbatim}
(1) property(possession, T2, Per, Obj) :- after(T2, T1),  
     property(possession, T1, Per, Obj), not neg_property(possession, T2, Per, Obj).
(2) neg_property(possession, T2, Per, Obj) :- after(T2, T1),   
     neg_property(possession, T1, Per, Obj), not property(possession, T2, Per, Obj).\end{verbatim}
            
            }
        \end{minipage}
        }

\subsection{Question translation}
Based on the natural language question, SQuARE generates the ASP query, which runs on the goal directed s(CASP) ASP engine to compute the answer. It mainly follows the ASP query generation mechanism described in \cite{aqua}. For a task, a general query rule is defined that is a collection of sub-queries needed to answer the question. This query rule can be thought of as part of commonsense knowledge. 
Unlike machine learning based systems, SQuARE is capable of handling all the datasets simultaneously,  with no loss in accuracy. Incremental and scalable design of SQuARE helps to use one reasoning component inside another, when needed. For example, the reasoning component to solve the \textit{single supporting fact} task becomes a sub-goal in the \textit{two supporting facts} task. 

To answer questions in the bAbI data sets, the systems also need to have commonsense understanding of various concepts. In the case of bAbI, commonsense knowledge of finding a location is needed, which is shown below:

\smallskip
\noindent
        \cprotect \fbox{
        \centering
        \begin{minipage}{0.97\linewidth}
            {
            \small
            \begin{verbatim}
(1) get_location(Per, Loc) :- get_all_times(Ts), get_location(Per, Loc, Ts).
(2) get_location(Per, Loc, Ts) :- filter_times(Per, Ts, Fil_Ts),
        get_max_time(Fil_Ts, MaxT), property(location, MaxT, Per, Loc).
(3) get_object_location(Obj, Loc) :- get_all_times(Ts),
        filtered_possession_times(Obj, Ts, Fil_Ts), get_max_time(Fil_Ts, MaxT),
        get_sublist_times(MaxT, Ts), property(possession, MaxT, Per, Obj),
        get_location(Per, Loc, Ts).\end{verbatim}
        }    
        \end{minipage}
        }

\smallskip
\noindent

The rules are self-explanatory. However, from a commonsense reasoning perspective, such rules can be thought of as a process template that a human may follow to achieve a specific task. Humans either learn these patterns on their own or someone imparts the knowledge to them. Supplying these rules assumes the latter.

Initially, the natural language questions are parsed using the Stanford's CoreNLP parser to get the parts-of-speech, the dependency graph, and the named-entities (if any). The named-entities help the system to identify that the query is about a person or a physical object. For example, if there are questions like - \textit{``Where is Sandra?''} and \textit{``Where is the football?''}, then named-entity recognizer identifies `Sandra' as a person. In the process of parsing, the system also detects the question type and the answer type, similar to the process described in \cite{caspr} and \cite{aqua}. For instance, \textit{`where'} wh-word normally asks for a location. Then the ASP query is formulated based on the information gathered by parsing the question. The queries for the questions - \textit{`where is Sandra?'} and \textit{`where is the football?'} are as follows respectively:   

\smallskip
\noindent
       \cprotect \fbox{
        \centering
        \begin{minipage}{0.95\linewidth}
            {\small
            \begin{verbatim}
?- get_location(sandra, Location).
?- get_object_location(the_football, Location).    \end{verbatim}
            
            }
        \end{minipage}
        }


\section{Example}\label{section:example}
To illustrate the capability of SQuARE further, we discuss a full-fledged example in this section showing the data-flow and the intermediate outputs from each step. 
Customized section of a story is taken from  Task-2.

\smallskip
\noindent    
\cprotect \fbox{
        \centering
        \begin{minipage}{0.95\linewidth}
            {
            \small
\begin{verbatim}
1 Mary moved to the bedroom.             4 Mary discarded the milk there. 
2 Mary got the milk there.               5 Mary journeyed to the bathroom.
3 Mary travelled to the hallway.
\end{verbatim}
            }
        \end{minipage}
    }
    
\smallskip
\noindent
The story tells that: \textit{Mary moved to the bedroom, got the milk and moved to the hallway, where she left the milk and moved to the bathroom}.

\noindent \textbf{Parsed Output: } The Stanford's CoreNLP parser generates the syntactic parse tree of each statement, that is next input to the semantic generation algorithm (Algorithm \ref{algorithm1}). Due to lack of space, only one parse tree is shown (statement 4 for story in Figure \ref{fig:parse-tree}).

\noindent \textbf{Matching VerbNet Frame: } Algorithm \ref{algorithm1} finds partial matching in the parse tree with the frame syntax from VerbNet and grounds the frame semantics. Figure \ref{fig:verbnet-example} illustrates an example of a partial matching frame for the verb \textit{discard} that is used in statement (4) of the story. 
%
It depicts that if the parse tree partially matches the syntactic structure - \textit{NP V NP PP} then the frame syntax variables ---  \textit{Agent, Theme, Destination}  will be grounded respectively with concrete objects from the sentence. These frame syntax variables are nothing but the thematic roles pre-defined in the VerbNet repository \cite{vn}. As we can see, in this scenario, there is a partial matching between the parse tree and the frame structure. The frame syntax variables are the predicate terms in the frame semantics.   

\noindent \textbf{Semantic Generation: } 
After processing the whole story and due to the partial matching, the program generates multiple semantics for different verbs. Because of space constraints, we show one of the many predicates of the instantiated semantics  of the verb \textit{discard} below. The process of semantic generation is discussed in Section \ref{section:semantic_algebra}. 

\noindent    
\cprotect \fbox{
        \centering
        \begin{minipage}{0.95\linewidth}
            {\small
            \verb|exert_force(t4, agent(mary), theme(the_milk)).|
            }

        \end{minipage}
    }
    
\vspace{0.02in}
\noindent
The variables \textit{Agent} and \textit{Theme} of the frame semantics (shown above) are grounded with the terms \textit{mary} and \textit{the milk}, respectively. Recall that frame semantics is coded in ASP.

\noindent \textbf{Question: } Now, the following natural language question is asked against the story. The actual answer is also given in the same line after the question. 

\noindent    
\cprotect \fbox{
        \centering
        \begin{minipage}{0.95\linewidth}
        {\small       \verb|Where is the milk?  	hallway|	            }
         
        \end{minipage}
    }

\vspace{0.02in}
\noindent \textbf{ASP Query: } Following is the ASP query along with the query rule for this type of question generated by our SQuARE system.
\vspace{0.02in}

\noindent    
\cprotect \fbox{
        \centering
        \begin{minipage}{0.96\linewidth}
        {\small
          \begin{verbatim}
get_object_location(Obj, Loc) :- get_all_times(Ts),
        filtered_possession_times(Obj, Ts, Fil_Ts), get_max_time(Fil_Ts, MaxT), 
        get_sublist_times(MaxT, Ts), property(possession, MaxT, Person, Obj),
        get_location(Person, Loc, Ts).
?- get_object_location(the_milk, Loc).\end{verbatim}  
        }      
        \end{minipage}
    }

\vspace{0.02in}
\noindent \textbf{Answer: } The answer found by the SQuARE system matches with the actual answer (\textit{hallway}).

\noindent \textbf{Justification:}
One major advantage of our framework is that it can provide a justification. This is possible because of the use of query-driven execution under the s(CASP) system.
The formatted justification tree for the query corresponding to \textit{``Where is the milk?''} is given below. The answer to the question is \textit{the hallway}, the second argument in the query \textit{get\_object\_location}. Providing a justification is as simple as printing the s(CASP) proof tree for the query (indentation represents call depth). In the future, we plan to work on converting this justification into English, similar to an explanation that a human would give. 

\noindent    
\cprotect \fbox{
        \centering
        \begin{minipage}{0.95\linewidth}

{
\small
\begin{verbatim}
BEGIN JUSTIFICATION
  get_object_location(the_milk, the_hallway)
    get_all_times([t1, t2, t3, t4, t5])
    filtered_possession_times(the_milk, [t1, t2, t3, t4, t5], [t2, t3])
    get_max_time([t2, t3], t3)
    get_sublist_times(t3, [t1, t2, t3])
    property(possession, t3, mary, the_milk)
    get_location(mary, the_hallway, [t1, t2, t3])
END JUSTIFICATION
\end{verbatim}
}
\end{minipage}
}


\section{Experiments and Results}

\noindent \textbf{The bAbI Dataset:} We have tested our SQuARE system on the bAbI dataset. The bAbI (pronounced as \textit{``baby''}) question answering tasks \cite{babi} were created by the Facebook AI research team to help NLU research. As the name indicates, the tasks are also very easy for humans  to reason about and answer with proper justification (though it should be noted that some stories have as many as 71 sentences, so the human effort can be non-trivial). The bAbI QA tasks comprise 20 different datasets for 20 independent reasoning tasks such as basic induction, deduction, temporal reasoning, chaining facts, and many more. 

This dataset was mainly created for train-and-test based deep learning models for natural language understanding. For each question asked after a set of declarative statements, the datasets provide the supporting facts for the actual answer as well. Not only it expects a learner to memorize the pattern in between a question and an answer, but also to grasp the supporting sentence for that answer. We mainly choose this dataset to show that a model can be created without training that can \textit{``truly understand''} the story and produce the correct answer to a question with proper justification. Moreover, another purpose of choosing the dataset is that it is actually composed of simple sentences and mainly focuses on reasoning. This helps us to concentrate more on knowledge representation and modeling than pre-processing/parsing of the English sentences.

\noindent \textbf{Tasks Description:} SQuARE was tested on five bAbI tasks: 
\begin{itemize}[noitemsep,topsep=0pt]
    \item \textit{\textbf{Single Supporting Fact:}} consists of simple questions that can be answered using single supporting fact given earlier in the story for that particular question. The other information given in the story is irrelevant.
    
    \item \textit{\textbf{Two and Three Supporting Facts:}} contains questions that require multi-hop reasoning to find the answer from two or three supporting facts that are chained together with  information possibly not relevant to that question.
    
    \item \textit{\textbf{Two Argument Relations:}} checks the ability of a system to recognize and differentiate the subject and object of a sentence and use commonsense knowledge to answer a question (e.g, \textit{if room A is south of room B, then B is north of A}). 
    \item \textit{\textbf{Positional Reasoning:}} tests the spatial reasoning of a system. A story describes a few object's positions and features (e.g., \textit{color of a box}), and asks questions about positional relationship between objects. ML-based system have the most difficulty in this task.
\end{itemize}

\setlength{\belowcaptionskip}{-14pt}
\begin{figure*}[b]
    \centering
    \includegraphics[width = 13cm]{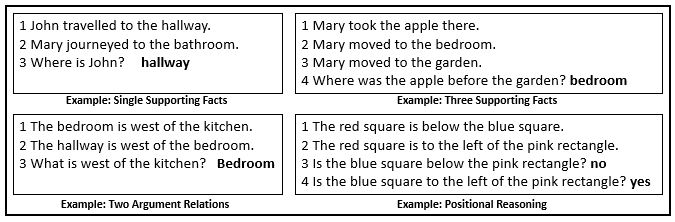}
    \caption{Examples of stories in various tasks}
    \label{story_examples}
    \end{figure*}

\noindent Example stories for four of the tasks are shown in Fig. \ref{story_examples}. Example story for the remaining task (\textit{two supporting facts}) was shown in Sec. \ref{section:example}. Note that the verb `\textit{is}' is not supported in VerbNet. Thus, commonsense knowledge describing the verb `\textit{is}', coded in ASP, had to be added.


\begin{table}[t]
\centering
\scriptsize
\setlength{\belowcaptionskip}{0.5pt}
\begin{tabular}{|l||*{5}{c|}}
\hline
\backslashbox[13em]{\hphantom{thequ}Tasks}{Attributes\hphantom{th}} & 
    \makebox[3em]{\textbf{\cellcolor{blue!25}\begin{tabular}[c]{@{}c@{}}No. of \\ Stories\end{tabular}}} &
    \makebox[6em]{\textbf{\cellcolor{blue!25}\begin{tabular}[c]{@{}c@{}}No. of \\Questions \\ per Story\end{tabular}}} &
    \makebox[9em]{\textbf{\cellcolor{blue!25}\begin{tabular}[c]{@{}c@{}}Avg. Story Size\\ (No. of Sentences)\end{tabular}}} & 
    \makebox[4em]{\textbf{\cellcolor{blue!25}\begin{tabular}[c]{@{}c@{}}Accuracy\\ (\%)\end{tabular}}} & 
    \makebox[6em]{\textbf{\cellcolor{blue!25}\begin{tabular}[c]{@{}c@{}}Avg. Time \\ per Question\\ (Seconds)\end{tabular}}}\\\hline\hline
\textbf{\cellcolor{green!20}Single Supporting Facts} & 2200 & 5 & 10 & 100 & 15.4  \\\hline
\textbf{\cellcolor{green!20}Two Supporting Facts} & 2200 & 5 & 22 & 100 & 42.8\\\hline
\textbf{\cellcolor{green!20}Three Supporting Facts} & 1800 & 5 & 31 & 100 & 147.1\\\hline
\textbf{\cellcolor{green!20}Two Argument Relations} & 11000 & 1 & 2 & 100 & 0.6\\\hline
\textbf{\cellcolor{green!20}Positional Reasoning} & 1375 & 8 & 2 & 100 & 0.3\\\hline
\end{tabular}
\vspace{0.02in}
\caption{Performance Results}
\label{table:performance_result}
\end{table}

\noindent \textbf{Experiment:} 
%
%
As discussed earlier, the stories are converted to an answer set program. The s(CASP) engine executes the translated ASP query to get the answer. In our experiments, this process is repeated for each story and for each question posed to that story. As SQuARE does not require any sort of training, we merged the training and testing datasets together to obtain one single repository for each task to test the SQuARE system. Table \ref{table:performance_result} summarizes our results. By achieving 100\% accuracy on all five datasets and providing the proof tree for each answer, SQuARE outperforms all the other machine learning and neural-net based systems in terms of accuracy and explainability. 

Average time to answer a question can range from 0.3 seconds to around 2.5 minutes. However, we are confident that these times can be greatly improved by careful analysis of the code based on execution profiling as well as improvements to s(CASP) engine that are continually being made. 
Note that, all the benchmarks are tested on a \textit{intel i9-9900 CPU} with \textit{16G RAM}.

\noindent \textbf{Comparison of Results \& Related Work:}
Table \ref{table:result_comparison} compares our result in terms of accuracy with other machine learning based models for the five bAbI datasets. N-gram classifier is the baseline model \cite{babi} for this dataset whereas the LSTM (long-short-term-memory recurrent neural network) \cite{lstm} is the most basic deep-learning model tested. The accuracies for both the N-gram classifier and the LSTM are the result of weak supervision used \cite{babi}, that means the models are trained only on the training dataset given. However, the Structured SVM \cite{babi}, dynamic memory network \cite{dmn}, and memory-neural-network (MemNN) with adaptive memory (AM), N-grams (NG), and non-linearity (NL) \cite{babi} are provided with external supporting facts for training. Multitasking with MemNN shows the accuracy if all the data from all the 20 tasks are used for training. For these tasks, in terms of accuracy, MemNN (AM, NG, NL) and SQuARE achieved 100\%. The SQuARE system can provide a justification for the answer, which other ML-based systems cannot.

\begin{table}[b]
\centering
\scriptsize
\setlength{\belowcaptionskip}{1pt}
\begin{tabular}{|c||*{5}{c|}}
\hline
\backslashbox[14em]{\hphantom{thequ}Models}{Tasks\hphantom{thequ}} & 
    \makebox[5em]{\textbf{\cellcolor{blue!25}\begin{tabular}[c]{@{}c@{}}Single \\ Supporting \\ Facts\end{tabular}}} & 
    \makebox[5em]{\textbf{\cellcolor{blue!25}\begin{tabular}[c]{@{}c@{}}Two \\ Supporting \\ Facts\end{tabular}}} & 
    \makebox[5em]{\textbf{\cellcolor{blue!25}\begin{tabular}[c]{@{}c@{}}Three \\ Supporting \\ Facts\end{tabular}}} & 
    \makebox[5em]{\textbf{\cellcolor{blue!25}\begin{tabular}[c]{@{}c@{}}Two \\ Arguments \\ Relations\end{tabular}}} &
    \makebox[5em]{\textbf{\cellcolor{blue!25}\begin{tabular}[c]{@{}c@{}}Positional \\ Reasoning\end{tabular}}}\\\hline\hline
\textbf{\cellcolor{green!20}N-gram Classifier} & 36 & 2 & 7 & 50 & 46 \\\hline
\textbf{\cellcolor{green!20}LSTM} & 50 & 20 & 20 & 61 & 51\\\hline
\textbf{\cellcolor{green!20}Structured SVM} & 99 & 74 & 17 & 98 & 61\\\hline
\textbf{\cellcolor{green!20}DMN} & 100 & 98.2 & 95.2 & 100 & 59.6\\\hline
\textbf{\cellcolor{green!20}Multi-Tasking (MemNN)} & 100 & 100 & 98 & 80 & 72\\\hline
\textbf{\cellcolor{green!20}MemNN (AM+NG+NL)} & 100 & 100 & 100 & 100 & 65 \\\hline
\textbf{\cellcolor{green!20}SQuARE} & 100 & 100 & 100 & 100 & 100\\\hline
\end{tabular}
\vspace{0.02in}
\caption{Accuracy (\%) comparison of SQuARE with other models for bAbI tasks}
\label{table:result_comparison}
\end{table}


ASP has also been applied to the bAbI dataset by Mitra and Baral \cite{mitra,mitrabaral18} and 
has been an inspiration for our work, 
however, theirs is still a machine learning-based system as it employs inductive LP and statistical methods along with ASP. The system achieves great accuracy, however, it requires annotated examples to be fed into the ILP system to derive the hypothesis. This process is iterative and requires ``manually finding the group size'' \cite{mitrabaral18}. Our SQuARE system, in contrast,  relies completely on automated reasoning. No manual intervention is required other than the necessary task of providing commonsense knowledge.
Also, in the ILP based approach, the `mode' declaration needs to be manually encoded for each verb used in each task, whereas SQuARE can generate ASP code automatically using our novel semantic knowledge generation algorithm (discussed in section \ref{section:semantic_algebra}).
Mitra and Baral's system learns the commonsense knowledge using ILP which requires manual annotation of the examples. In contrast, (reusable) commonsense knowledge has to be hand-coded in the SQuARE system. In that sense, one could argue that Mitra and Baral's approach is better as commonsense knowledge is produced automatically. In their case, however, generation of knowledge from the scenario expressed in  natural language \textit{also} requires manual annotation. That part is completely automated in our approach, thanks to our partial matching algorithm and VerbNet. Thus, the SQuARE system further advances state of the art in automated reasoning based approaches for QA. 

With respect to commonsense knowledge, we take a different approach: we want to mimic a human in that all commonsense knowledge that an average human possesses should be coded in ASP using predefined simple generic predicates (such as {\tt property, cause, after,} etc.) and made available to the question answering system. Use of generic predicates  simplifies the development of commonsense knowledge, making it reusable and resulting in a scalable QA system. Of course, the amount of knowledge to be coded may be daunting, but it can be constructed incrementally on an as-needed basis. The amount of work would not be any less than annotating examples and generating commonsense knowledge automatically using a machine learning algorithm such as ILP. There may be additional work needed as the commonsense knowledge learned using ILP may have to be manually reviewed for errors and omissions and then manually fixed.  Also, the generated commonsense knowledge should follow a standard naming convention for predicates, as we discussed earlier, something that is harder to realize in the ILP approach. If standardized conventions are not followed, the generated commonsense knowledge cannot be of any use to other applications without re-annotation and re-training. 

An action language based QA methodology using VerbNet has been developed by Lierler et al \cite{ALM}. The project aims to extend frame semantics with ALM, \textit{an action language} \cite{ALM}, to provide interpretable semantic annotations. Unlike SQuARE, it is not an end-to-end automated QA system. 

Cyc \cite{cyc} is one of the earliest AI projects that aims to capture commonsense knowledge and reason over it. In Cyc, the acquired commonsense knowledge about the world is represented in the form of a vast collection of ontologies that consist of basic concepts and implicit rules about how the world works. To solve an inference problem, Cyc uses multiple reasoning agents in collaboration and these agents rely on more than 1000 pre-modeled heuristic modules. This is one of the challenges of Cyc as it is very hard to decide which model to apply and when, and which heuristic to use. A commonsense reasoning system should be designed in a simpler way, otherwise we may have acquired all the individual pieces of knowledge to answer and query, but may not be able to compose the answer from the pieces. For this reason, SQuARE uses minimal number of generic predicates and simple ASP-based reasoning rules (default reasoning is used as much as possible). Cyc did not support natural language question answering, however, recently some efforts have been initiated in that direction.

\section{Discussion}
Our goal is to create a semantic-based textual QA framework that mimics the way humans answer questions. Humans understand a passage semantics and use commonsense knowledge to logically reason to find answers. We believe that this process is the most effective way to develop an automated QA system. Intelligent behavior in humans is due to both learning and reasoning. At present, machine learning dominates AI. Machine learning is useful for many tasks (e.g., in parsing English sentences, detecting sentiment, etc.), however, it is not sufficient by itself for completely automating tasks that require intelligence. For automating intelligent behavior we need automated commonsense reasoning as well. The SQuARE system is a step in that direction. Note that it does not require any manual coding other than providing commonsense knowledge. 


The SQuARE system has many advantages over a machine learning based  QA systems. It can answer questions without training. For example, none of the ML-based systems discussed in Table \ref{table:result_comparison} will be able to answer the question - \textit{``Is milk a liquid?''} without appropriate training data. SQuARE can answer such questions based on its commonsense knowledge, just like a human. SQuARE is also capable of providing explanation for each computed answer. Providing explanation shows a system's true ``understanding'' of the matter, which is a necessary feature of a truly intelligent system. Interpretability makes SQuARE transparent, that means the system can be well understood, debugged, and improved. Also, our approach is incremental in nature. The system can be easily expanded to cover more NLU tasks without hampering the earlier accuracy. On the contrary, expanding the capabilities of a machine learning system often requires hyper-parameter tuning, which often results in reduced accuracy. 

\section{Future Work and Conclusions}

This paper proposes a semantic based general question answering framework that uses commonsense knowledge to generate an answer with proper justification. This robust framework for automated QA preserves scalability, interpretability and explainability.  We also demonstrated an end-to-end QA system called SQuARE that uses this framework to convert  textual knowledge to semantic representation and reason over it with goal-directed ASP to answer questions posed in English.  SQuARE  relies on the s(CASP) goal-directed system and produced 100\% accuracy on five bAbI tasks on which it was tested. The SQuARE system is a step towards general QA systems that can answer natural language questions against any text, automated conversational AI systems, etc.

As part of our future work, we plan to complete the remaining 15 bAbI tasks. The bAbI suite has many other more advanced tasks as well (\textit{understanding children's books, understanding movie dialogs,} etc.). We plan to extend SQuARE to implement those tasks too. Eventually, our goal is to develop a conversational AI system based on automated commonsense reasoning (using the goal-directed s(CASP) system) that can ``converse'' with a person based on ``truly understanding'' that person's dialogs. 
We also plan to automate the commonsense knowledge generation process using state-of-the-art ILP systems \cite{fold}.



\end{document}